\documentclass{article}

\usepackage{microtype}
\usepackage{graphicx}
\usepackage{subcaption}
\usepackage{booktabs} 

\usepackage{hyperref}

\usepackage[preprint]{icml2026}

\usepackage{amsmath}
\usepackage{amssymb}
\usepackage{mathtools}
\usepackage{amsthm}

\usepackage[capitalize,noabbrev]{cleveref}

\theoremstyle{plain}

\theoremstyle{definition}

\theoremstyle{remark}

\usepackage[textsize=tiny]{todonotes}

\icmltitlerunning{SynHLMA: Discrete HAOI Representation for Articulated Object Manipulation}

\begin{document}

\twocolumn[
  \icmltitle{SynHLMA: Synthesizing Hand Language Manipulation for Articulated Object with Discrete Human Object Interaction Representation}



\begin{icmlauthorlist}
\icmlauthor{Zhi Wang}{equal,hfut}
\icmlauthor{Yuyan Liu}{equal,hfut}
\icmlauthor{Liu Liu}{hfut}
\icmlauthor{Li Zhang}{ustc}
\icmlauthor{Ruixuan Lu}{ahu}
\icmlauthor{Dan Guo}{hfut}
\end{icmlauthorlist}

\icmlaffiliation{hfut}{Hefei University of Technology, Hefei, China}
\icmlaffiliation{ustc}{University of Science and Technology of China, Hefei, China}
\icmlaffiliation{ahu}{Anhui University, Hefei, China}

\icmlcorrespondingauthor{Liu Liu}{liuliu@hfut.edu.cn}

  \icmlkeywords{Machine Learning, ICML}

  \vskip 0.3in
]



\printAffiliationsAndNotice{\icmlEqualContribution}

\begin{abstract}
Generating language-guided hand grasps has been widely studied in embodied AI. 
However, extending grasp synthesis to \textbf{H}and \textbf{A}rticulated \textbf{O}bject \textbf{I}nteraction (\textbf{HAOI}) requires modeling not only object functionality but also temporally coherent manipulation along object articulation. 
This paper proposes a novel HAOI sequence generation framework \textbf{SynHLMA}, to \textbf{Syn}thesize \textbf{H}and \textbf{L}anguage \textbf{M}anipulation for \textbf{A}rticulated objects.
Given an articulated object point cloud and a textual instruction, SynHLMA learns a hierarchical discrete representation of hand–object interaction frames and aligns them with language embeddings through a manipulation language model in a shared semantic space. 
An \textbf{articulation-aware objective} further enforces geometric validity and joint-consistent temporal dynamics, enabling physically grounded long-horizon generation.
SynHLMA supports HAOI generation, prediction, and interpolation within a single autoregressive formulation. 
We evaluate our method on the newly constructed \textbf{HAOI-Lang} dataset and demonstrate superior performance over state-of-the-art baselines. 
Moreover, the generated manipulation sequences effectively facilitate imitation learning for dexterous robotic grasp execution. 
Code and dataset will be publicly released.
\vspace{-6mm}
\end{abstract}

\begin{figure}[ht]
  \includegraphics[width=0.9\linewidth]{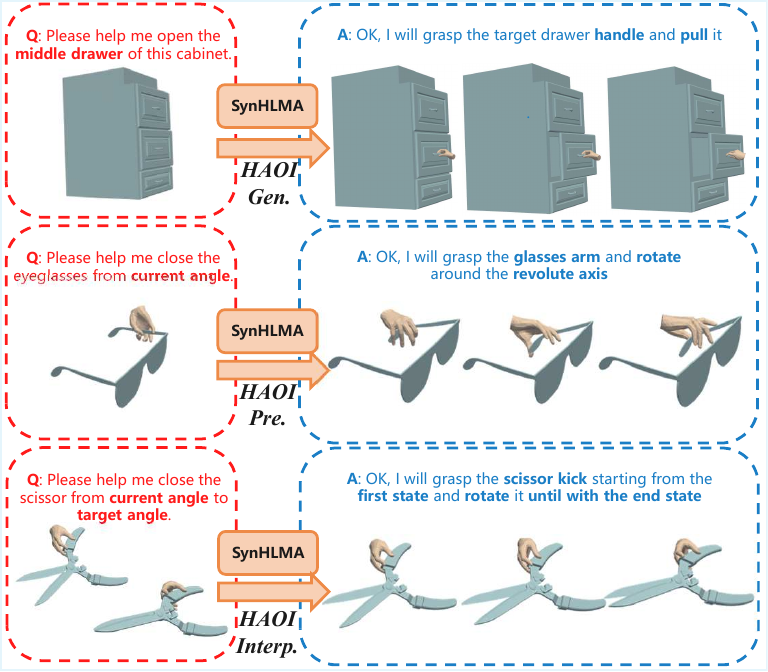}
  \centering
  \caption{Given an articulated object shape and a natural language as instruction, our SynHLMA achieves three typical hand manipulation synthesis tasks: HAOI generation (up), HAOI prediction (middle) and HAOI interpolation (down) with their corresponding manipulation descriptions.}
  \label{fig:start}
  \vspace{-8mm}
\end{figure}  

\section{Introduction}
Understanding human manipulation intent in human–object interaction (HOI) remains a fundamental challenge for robotic dexterity. 
While recent advances have enabled physically plausible grasp synthesis for rigid objects, articulated objects require modeling not only stable grasp poses but also temporally coherent deformation processes. 
For example, using scissors entails both grasp acquisition and coordinated articulation control (e.g., opening and closing). 
Such interactions demand reasoning over evolving object configurations and state-dependent affordances. 
We refer to this problem as \textbf{H}uman \textbf{A}rticulated \textbf{O}bject \textbf{I}nteraction (\textbf{HAOI}) generation under language-guided intent.

Despite substantial progress in grasp synthesis, existing approaches exhibit notable limitations. 
Robotic-hand-based methods~\cite{jin2024reasoning,xie2023learning} often lack human-hand realism, while skeleton-driven techniques~\cite{yang2022artiboost,karunratanakul2021skeleton} neglect physically grounded contact modeling. 
Contact-centric approaches~\cite{grady2021contactopt,yang2022oakink} struggle to integrate language semantics with articulated object dynamics. 
Moreover, diffusion-based models suffer from degraded long-horizon generation due to weak structural priors, and most prior work focuses narrowly on text-to-grasp tasks without modeling full manipulation sequences.

Articulated objects introduce intrinsic structural variability into HOI. 
Unlike rigid manipulation—where contact topology and kinematic constraints remain largely stable—articulated interactions exhibit configuration-dependent contacts, joint-coupled motion dependencies, and temporally evolving affordances. 
These properties frequently induce inconsistencies between hand motion and object deformation, significantly increasing the difficulty of coherent sequence generation.

To address these challenges, we propose \textbf{SynHLMA}, a unified framework for \textbf{Syn}thesizing \textbf{H}and–\textbf{L}anguage \textbf{M}anipulation of \textbf{A}rticulated objects. 
Our key insight is that articulated manipulation exhibits discrete structural regularities analogous to linguistic tokens and grasp taxonomies~\cite{feix2015grasp}. 
We therefore introduce a hierarchical discrete representation for HAOI. 
Specifically, two modular VQ-VAE models encode object articulation states, global hand configuration, local pose parameters, and refinement residuals into structured manipulation tokens. 
This design enables semantic alignment between articulation states and grasp configurations, promoting coordinated hand–object dynamics across varying object states.

Building upon the discrete representation, we develop a manipulation language model that aligns tokenized HAOI sequences with language embeddings via LoRA adaptation in a shared semantic space. 
An autoregressive formulation predicts incremental state differences, improving long-horizon stability and enabling three tasks: HAOI generation, prediction, and interpolation (Fig.~\ref{fig:start}).

A central challenge in articulated manipulation modeling lies in preserving articulation consistency across geometry, kinematics, and temporal evolution. 
Existing approaches typically optimize either continuous pose reconstruction or discrete token likelihood, without explicitly enforcing articulation constraints. 
Consequently, generated sequences may exhibit interpenetration, inconsistent joint states, or temporally incoherent motion.

To overcome this limitation, we introduce an \emph{articulation-aware objective} that integrates multi-level structural constraints into training. 
Specifically, our objective enforces: 
(1) geometric validity to prevent hand–object penetration and ensure accurate joint decoding; 
(2) hierarchical reconstruction consistency between discrete tokens and continuous kinematics; 
(3) temporal articulation coherence across sequential states. 
By embedding articulation awareness directly into representation learning and sequence modeling, SynHLMA generates physically grounded and semantically consistent manipulation trajectories.

To support this task, we construct \textbf{HAOI-Lang}, a large-scale dataset of articulated manipulation sequences with language annotations. 
Using a physics-based interaction engine, we sample diverse contact points and articulation trajectories for each object. 
GPT-4 is employed to generate rich textual descriptions covering intent, direction, and spatial relations, followed by human refinement to ensure semantic fidelity. 
Experiments demonstrate that SynHLMA consistently outperforms prior baselines on HAOI generation, prediction, and interpolation, and that the synthesized sequences effectively facilitate imitation learning for dexterous robotic hands.

In summary, our work makes the following key contributions:
1) \textbf{Discrete Manipulation Representation.} We propose a hierarchical tokenization scheme for articulated manipulation, enabling structured and controllable sequence generation.
2) \textbf{Manipulation Language Model.} We present a language-aligned generative model for articulated object manipulation supporting generation, prediction, and interpolation.
3) \textbf{Articulation-Aware Objective.} We introduce a unified training objective that enforces geometric validity, joint-state alignment, and temporal coherence.
4) \textbf{HAOI-Lang Dataset.} We construct a new language-annotated dataset for articulated hand–object interaction.


\begin{figure*}[htb]
  \centering
  \includegraphics[width=0.8\linewidth]{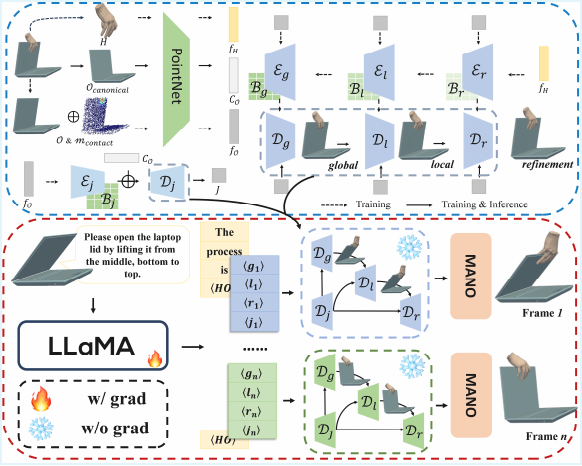}
  \caption{Our SynHLMA pipeline, the upper blue dashed region illustrates the training process of our proposed Discrete Articulated Manipulation Representation model. The lower red region depicts the HAOI Manipulation Language Model, where the parameters of the Discrete Articulated Manipulation Representation are kept frozen during training.}
  \label{fig:pipline}
  \vspace{-4mm}
\end{figure*} 

\section{Related Works}
\subsection{Hand Grasp Synthesis}
Grasp synthesis for hand poses has a long research history. Following current trends, existing methods can be broadly categorized into two main approaches:  probabilistic grasp generation and simulator-based grasp generation.

Probabilistic grasp generation can be broadly categorized into two approaches: regression-based and diffusion-based methods. In regression-based models \cite{li2024semgrasp,huang2025hoigpt,prokudin2019efficient,yang2022oakink,yang2021cpf}, a point cloud encoder is first used to extract features representing the HOI. The model then learns the distribution of these features and samples from it, followed by a regression module that predicts the hand parameters to complete the grasp generation.
In contrast, diffusion-based methods \cite{ye2023affordance,zhang2024nl2contact} gradually add noise to existing HOI data and learn to reverse this process through denoising. By modeling the denoising trajectory, the model captures the underlying distribution of HOI configurations in the dataset. Grasp generation is then achieved by iteratively denoising a noise-initialized sample.

Although probabilistic generation methods have become increasingly mature and object datasets \cite{deitke2023objaverse,mo2019partnet,liu2022akb} are steadily improving, there remains a significant lack of high-quality HOI datasets. With the advancement of virtual physics engines \cite{hussain2020unity,hwangbo2018per,xiang2020sapien}, simulator-based grasp generation has begun to attract growing attention. For example, \citet{xu2023unidexgrasp,zhang2024graspxl,christen2022d} proposed a method that employs a reinforcement learning framework to learn grasping reward strategies, with reward signals provided by a physics engine. This approach enables the large-scale generation of HOI datasets.

\subsection{Multimodal Large Language Models}
In recent years, the remarkable success of large language models (LLMs) such as GPT \cite{achiam2023gpt}, Qwen \cite{bai2023qwen}, Gemini \cite{team2023gemini} and Gemma \cite{team2024gemma} in various downstream natural language tasks—including text translation \cite{devlin2019bert}, semantic understanding \cite{du2022glam}, and text generation \cite{zhang2020pegasus} has inspired advancements in other domains. Many researchers have attempted to integrate LLMs with multimodal data, including audio \cite{borsos2023audiolm}, images \cite{radford2021learning}, videos \cite{li2023videochat}, and 3D object models \cite{jiang2023motiongpt}, to develop powerful multimodal models.

\citet{jiang2023motiongpt,guo2022tm2t,huang2025hoigpt,zhang2023generating} discretize poses and motions into tokens via encoders, then leverage pretrained large language models (LLMs) to achieve multimodal alignment for motion generation. \citet{cha2024text2hoi} aligns object and text features through PointNet and CLIP models respectively, and employs a diffusion model to learn feature distributions for generation. In our work, we adopt a similar paradigm and fine-tune Vicuna to generate grasping sequences conditioned on hand-object interaction features and natural language instructions. 

\section{Method}
\subsection{Overview}
We propose SynHLMA, a unified framework for generating HAOI sequences. SynHLMA employs two modular VQ-VAE models to discretize articulated object joint trajectories and hand grasps into structured tokens, upon which a manipulation language model learns action distributions conditioned on language instructions. To ensure physically plausible and structurally consistent generation, we introduce an articulation-aware objective that enforces geometric validity, joint-state alignment, and temporal coherence. Furthermore, we construct a large-scale physics-based simulation dataset with GPT-assisted annotations and manual refinement.

\subsection{Discrete Articulated Manipulation Representation}
\label{sec:tokengrasp}

We represent a hand grasp configuration as
$\boldsymbol{\beta} = (R, P, T),$
where $R \in \mathbb{R}^{3}$ denotes the axis–angle representation of the global hand rotation, 
$P \in \mathbb{R}^{90}$ represents the articulated hand pose parameters, and 
$T \in \mathbb{R}^{3}$ denotes the global hand translation. 
Given $\boldsymbol{\beta}$, the MANO parametric hand model $\mathcal{M}(\cdot)$ produces the 
corresponding hand mesh $\mathcal{H} \in \mathbb{R}^{778 \times 3}$:$\mathbf{\mathcal{H}} = \mathcal{M}(\boldsymbol{\beta}) = \mathcal{M}({R}, {P}, {T})$.

Unlike rigid-object manipulation, articulated objects introduce configuration-dependent 
geometric and kinematic variations, which require explicit modeling of joint states 
and their interaction with hand motion. To achieve semantically structured 
and articulation-consistent representations, we discretize both object articulation 
and hand manipulation into tokenized latent spaces using two modular VQ-VAE models. 
(Fig.~\ref{fig:pipline}).

We first discretize the articulated object state. 
Let $J \in \mathbb{R}^{3}$ denote the object joint parameters 
(e.g., rotational or translational state). 
We employ a single-layer VQ-VAE with encoder $\mathcal{E}_j$, 
decoder $\mathcal{D}_j$, and codebook $\mathcal{B}_j$ 
to quantize $J$ into a discrete token $\langle j \rangle$, where $j \in \mathbb{N}$. 
This mapping converts continuous joint variations into semantically meaningful indices, 
from which the original joint parameters can be reconstructed. 
By explicitly tokenizing articulation states, we establish a compact and structured 
representation of object configuration that can condition downstream manipulation generation.

To model the structured nature of human grasping, we decompose manipulation into 
three hierarchical components:
$\langle g, l, r \rangle,$
corresponding to global hand configuration, local pose articulation, 
and refinement residuals, respectively. 
Each component is quantized using a dedicated VQ-VAE branch with 
codebook $\mathcal{B}_i$, encoder $\mathcal{E}_i$, and decoder $\mathcal{D}_i$, 
where $i \in \{g, l, r\}$. 

This hierarchical design enables coarse-to-fine discretization of grasp parameters, 
encouraging semantic separation between global motion, articulated finger pose, 
and fine-grained refinement. The joint token $\langle j \rangle$ is incorporated 
as conditioning information to ensure articulation-aware manipulation modeling.

The decoding process mirrors the hierarchical structure and follows a 
coarse-to-fine conditional factorization.
First, the global hand transformation is generated conditioned on the 
global token and the object articulation token:
$\hat{R}, \hat{T} = \mathcal{D}_{g}(g, j).$ Next, the articulated hand pose is predicted conditioned on 
the global context and articulation state:
$\hat{P} = \mathcal{D}_{l}(g, l, j).$ Finally, a refinement stage predicts residual offsets conditioned on 
the full token set:
$\hat{\Delta R}, \hat{\Delta P}, \hat{\Delta T}
= \mathcal{D}_{r}(g, l, r, j).$ The final grasp parameters are obtained via residual composition:
$\hat{\boldsymbol{\beta}} =
\left(
\hat{\Delta R} \cdot \hat{R}, \;
\hat{\Delta P} + \hat{P}, \;
\hat{\Delta T} + \hat{T}
\right).$
 

This structured decoding explicitly models hierarchical token dependencies, allowing articulation states to influence global motion, articulated pose, and fine-grained refinement in a unified manner. As a result, the discrete representation captures object articulation dynamics and coordinated hand manipulation within a semantically aligned latent space.

\begin{figure}[ht]
  \begin{center}
    \centerline{\includegraphics[width=0.8\linewidth]{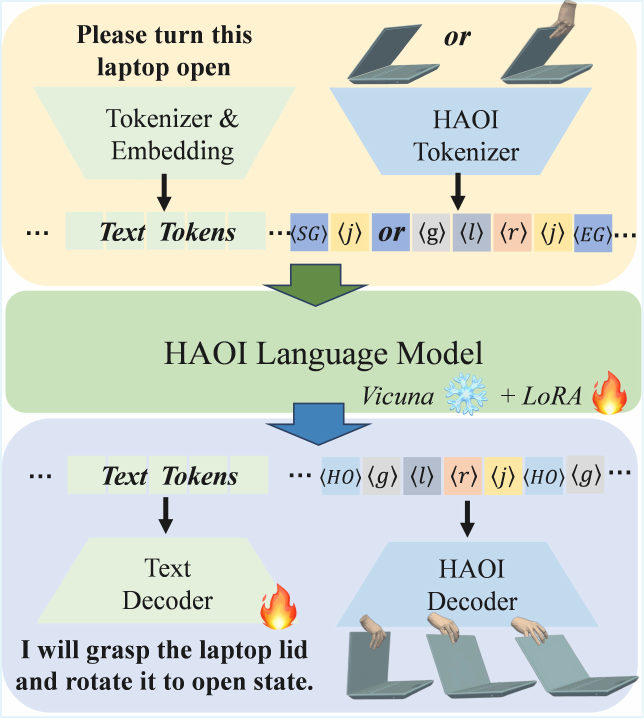}}
    \caption{
     HAOI Manipulation Language Model
    }
    \label{fig:MLLM}
  \end{center}
  \vspace{-8mm}
\end{figure}

\label{sec:multimodalgrasp}
\subsection{HAOI Manipulation Language Model}
Benefiting from the previously introduced Discrete Articulated Manipulation Representation, we further develop a HAOI Manipulation Language Model. As illustrated in Fig. \ref{fig:MLLM}, our model is designed to align three heterogeneous modalities: human manipulation, object features, and language instructions. 

After training the VQ-VAE, its parameters are frozen. We then utilize the VQ-VAE encoder $\mathcal{E}$ to obtain discrete tokens from HAOI sequences. The specific tokens provided to the MLLM vary depending on the downstream task. For the HAOI generation task, the objective is to predict subsequent grasps conditioned on the current object state; therefore, only the current object joint token $\langle j \rangle$ is required as input. In contrast, the HAOI prediction and interpolation tasks depend on the complete state of the current or intermediate interaction, and thus require the full token sequence $\langle g \rangle \langle l \rangle \langle r \rangle \langle j \rangle$ as input. To distinguish manipulation tokens from textual instructions, special markers \texttt{<SG>} and \texttt{<EG>} are appended to the token sequence to explicitly denote its beginning and end. 

Regardless of the downstream task formulation, the generation objective of the MLLM is to produce a complete HAOI sequence, namely $\langle g \rangle \langle l \rangle \langle r \rangle \langle j \rangle$. To further differentiate generated manipulation tokens from natural language tokens, we introduce special tokens \texttt{<HO>} to mark the start and end of manipulation generation. 
The overall generated grasp sequence can thus be expressed as follows: 
\[ \small
\langle HO \rangle \langle g_1 \rangle \langle l_1 \rangle \langle r_1 \rangle \langle j_1 \rangle \cdots \langle g_t \rangle \langle l_t \rangle \langle r_t \rangle \langle j_t \rangle \langle HO \rangle,
\]
where \(t\) is the total number of steps. For HAOI manipulation generation, \(\langle HO \rangle\) tokens are removed, and each tuple \(\langle g,l,r,j \rangle\) is sequentially mapped through the pre-trained multi-layer VQ-VAE codebook to obtain latent vectors. These vectors are then decoded by the VQ-VAE decoder $\mathcal{D}$ to produce the grasp configuration for each frame.

Multimodal features are projected into a unified semantic space via a linear projection and embedding layer. Denoting all inputs as \(S_{\text{input}}\) and outputs as \(S_{\text{out}} = \{ x_i \}_{i=1}^{t} \in \mathbb{R}^{t}\), sequence generation is formulated as a next-token prediction problem:
{\small
\begin{equation}
P(x_{i} | x_{<i}) = P(x_{i} | x_{1}, x_{2}, \ldots, x_{i-1}), i = 1, 2,...,t.
\end{equation}}

The HAOI manipulation language model is fine-tuned in two stages:
1) \textbf{Multimodal Alignmen}t: Embed new special tokens and align original embeddings with discrete manipulation representations.
2) \textbf{Instruction Tuning}: Freeze Stage 1 embeddings and tokenizer; optimize joint generation of grasp sequences and language outputs for stable, consistent behavior.

\subsection{Articulation-Aware Training Objective}
We train the proposed framework in two stages: (1) learning discrete articulated manipulation representations via VQ-VAE, and (2) fine-tuning the HAOI manipulation language model. The overall objective integrates geometry-aware regularization, hierarchical reconstruction, vector-quantization commitment, and sequence modeling losses.

\textbf{Geometry-Aware Regularization.}
To ensure physically plausible hand–object interactions and accurate joint-state modeling, we introduce two geometric constraints.

The penetration loss penalizes interpenetration between the predicted hand mesh $\hat{\mathcal{H}}$ and the articulated object mesh $\mathcal{O}$. Let $V$ denote the vertices of $\hat{\mathcal{H}}$, and $V_{\text{in}} \subseteq V$ be the subset penetrating the object surface. For each $v_i \in V_{\text{in}}$, we compute the squared distance to its closest surface point $v_i^{\mathcal{O}}$:
{\small
\begin{equation}
\mathcal{L}_{\text{pen}} 
= 
\frac{1}{|V_{\text{in}}|}
\sum_{v_i \in V_{\text{in}}}
\left\| v_i - v_i^{\mathcal{O}} \right\|_2^2.
\end{equation}}

For the articulated object branch, the joint reconstruction loss ensures accurate recovery of the joint state $J$:
{\small
\begin{equation}
\mathcal{L}_{\text{joint}}
=
\left\|
\mathcal{D}_{j}(\hat{z}_j)
-
J_{\text{gt}}
\right\|_2^2,
\end{equation}}
where $\hat{z}_j$ denotes the quantized latent representation of the joint branch.
The geometry-aware regularization is thus:
{\small
\begin{equation}
\mathcal{L}_{\text{geom}}
=
\lambda_{\text{pen}} \mathcal{L}_{\text{pen}}
+
\lambda_{\text{joint}} \mathcal{L}_{\text{joint}}.
\end{equation}}

\textbf{Hierarchical Reconstruction Loss.}
For the multi-stage VQ-VAE modeling hand manipulation, reconstruction is performed at three hierarchical levels corresponding to the global ($g$), local ($l$), and refinement ($r$) branches. Let $\mathcal{M}(R,P,T)$ denote the MANO mapping from grasp parameters to mesh vertices. The three levels are unified into a single reconstruction objective:
\vspace{-1.5mm}
\begin{equation}
\resizebox{0.9\linewidth}{!}{$
\begin{aligned}
\mathcal{L}_{\text{rec}} 
=&\;
\lambda_g
\left\|
\mathcal{M}(R,0,T)
-
\mathcal{M}(\hat{R},0,\hat{T})
\right\|_2^2 \\
&+
\lambda_l
\left\|
\mathcal{M}(R,P,T)
-
\mathcal{M}(\hat{R},\hat{P},\hat{T})
\right\|_2^2 \\
&+
\lambda_r
\left\|
\mathcal{M}(R,P,T)
-
\mathcal{M}(\Delta \hat{R} \cdot \hat{R},
\Delta \hat{P} + \hat{P},
\Delta \hat{T} + \hat{T})
\right\|_2^2
\end{aligned}
$}
\end{equation}
\vspace{-4mm}

The first term supervises coarse global pose prediction, the second term refines articulated hand configuration, and the third term enforces residual correction for fine-grained alignment.

\textbf{VQ-VAE Commitment Loss.}
For each branch $i \in \{g,l,r,j\}$, the encoder output $z_i$ is quantized to its nearest codebook embedding $\hat{z}_i$. The commitment loss stabilizes codebook learning and prevents encoder drift:
{\small
\begin{equation}
\mathcal{L}_{\text{com}}
=
\sum_{i \in \{g,l,r,j\}}
\left(
\left\| \mathrm{sg}[z_i] - \hat{z}_i \right\|_2^2
+
\beta
\left\| z_i - \mathrm{sg}[\hat{z}_i] \right\|_2^2
\right),
\end{equation}}
where $\mathrm{sg}[\cdot]$ denotes the stop-gradient operator.

\textbf{Manipulation Language Modeling Loss.}
For the HAOI manipulation language model, sequence generation is formulated as next-token prediction. Given token sequence $\{x_i\}_{i=1}^{t}$, the negative log-likelihood loss is:
{\small
\begin{equation}
\mathcal{L}_{\text{NLL}}
=
-
\sum_{i=1}^{t}
\log P(x_i \mid x_{<i}).
\end{equation}}

To enforce temporal consistency between adjacent frames, we introduce a pose consistency loss:
\begin{equation}
\resizebox{0.9\linewidth}{!}{$
\mathcal{L}_{\text{temp}}
=
\begin{cases}
\sum_{i=1}^{n}
\arccos
\left(
\frac{
\mathrm{Tr}\!\left(
\Delta \mathcal{R}_{J_i}
\Delta \mathcal{R}_{R_i}^{\top}
\right)
-1
}{2}
\right),
& \text{rotational joint,} \\[6pt]
\sum_{i=1}^{n}
\|
\Delta J_i - \Delta T_i
\|_2,
& \text{translational joint.}
\end{cases}
$}
\end{equation}

Based on the decoding procedure described in Section~\ref{sec:tokengrasp}, the output tokens of the manipulation language model are converted into the joint axis state $J$ for each frame. For rotational joints, $J$ encodes the object's axis--angle representation. We convert $J$ into a rotation matrix $\mathcal{R}_J$ using the Rodrigues formula, and compute the inter-frame rotational change as $\Delta \mathcal{R}_{J_i} = \mathcal{R}_{J_i} \mathcal{R}_{J_{i-1}}^\top$. The hand's rotational change $\Delta \mathcal{R}_{R_i}$ is obtained analogously. For translational joints, $J$ directly represents the object's translation vector; thus, $\Delta J_i$ and $\Delta T_i$ denote the inter-frame displacement of the object and the hand, respectively. Here, $n$ denotes the number of inter-frame transitions, excluding the first frame which has no predecessor.

\textbf{Overall Objective.}
The final training objective jointly minimizes:
{\small
\begin{equation}
\mathcal{L}
=
\mathcal{L}_{\text{geom}}
+
\mathcal{L}_{\text{rec}}
+
\mathcal{L}_{\text{com}}
+
\mathcal{L}_{\text{NLL}}
+
\lambda_{\text{temp}} \mathcal{L}_{\text{temp}}.
\end{equation}}
We collectively refer to the above formulation as an articulation-aware learning objective, 
as it explicitly enforces geometric validity, joint-state correctness, hierarchical grasp 
reconstruction, and temporal articulation consistency.

\subsection{HAOI-Lang Dataset Construction}
\label{sec:datagen}

Existing datasets lack articulated object manipulation sequences paired with language descriptions. A comprehensive joint-body dataset should satisfy: \textbf{Completeness} (covering common real-world joint types), \textbf{Physical Plausibility} (grasps obey physical and kinematic constraints), \textbf{Generality} (including small and large objects), and \textbf{Consistency} (object motions and corresponding language are standardized and unambiguous). To address this, we construct \textbf{HAOI-Lang}, a novel dataset of HAOI manipulation sequences guided by language instructions, built upon the ArtImage dataset~\cite{xue2021omadobjectmodelarticulated} within the PartNet-Mobility repository~\cite{xiang2020sapien}, and leverage a physics simulator to emulate realistic manipulation scenarios.

\begin{figure}[ht]
\centering
  \begin{center}
    \centerline{\includegraphics[width=0.9\columnwidth]{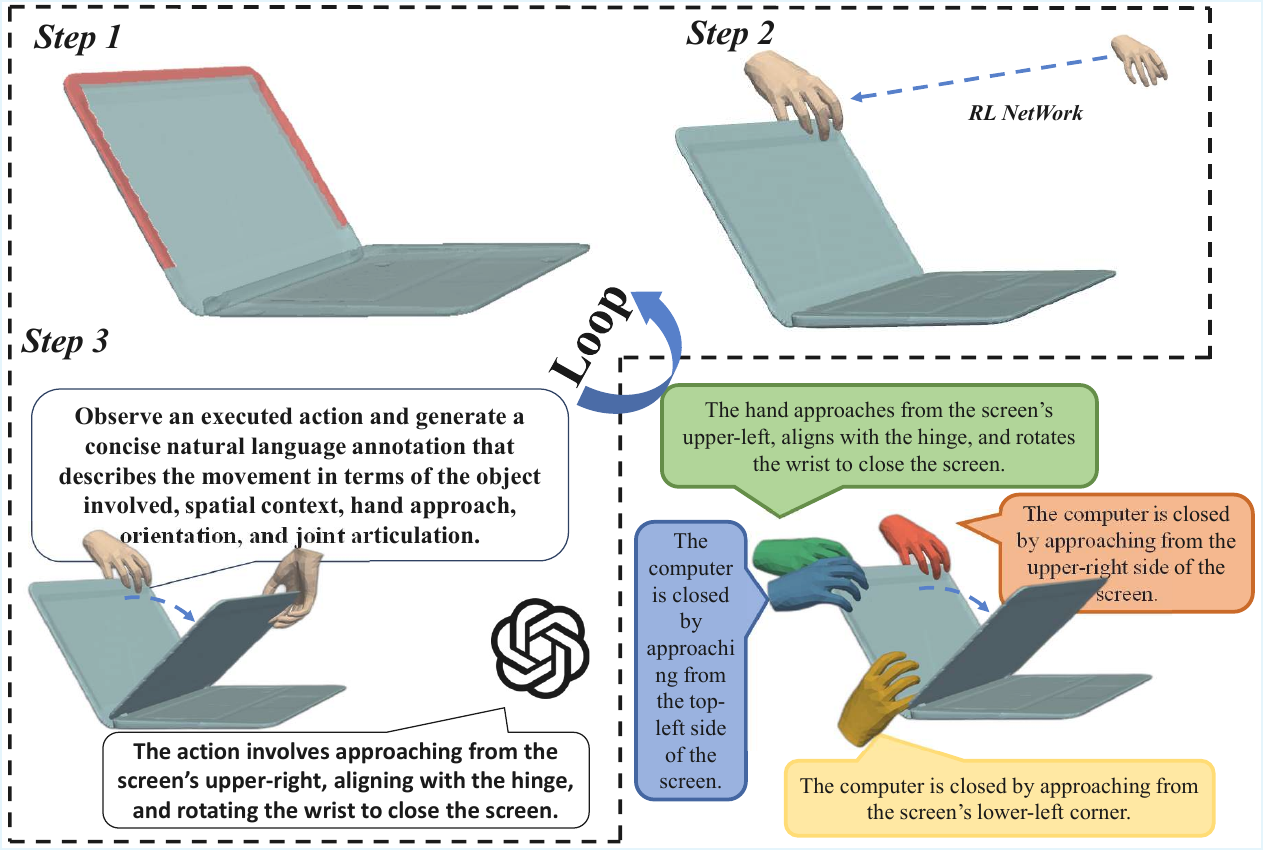}}
    \caption{
     Synthesis HAOI data generation process. (a) grasp region perception. (b) manipulation sequence generation with RL. (c) language instruction generation with GPT-4. (d) large number of samples demonstration.
    }
    \label{fig:graspxlgen}
  \end{center}
  \vspace{-4mm}
\end{figure}

We employ \textbf{RaiSim} as the simulation engine, which stands out for its use of the efficient \textit{Articulated Body Algorithm (ABA)} for forward dynamics. By avoiding explicit mass matrix computation and directly solving for generalized accelerations, ABA achieves $\mathcal{O}(n)$ complexity—particularly beneficial for high-DoF systems.

For grasping, we follow the reinforcement learning approach of~\citet{zhang2024graspxl}, which unifies hand–object grasp planning across diverse motion objectives, object geometries, and hand morphologies. The method segments objects into graspable and non-graspable regions and uses a reward function \(r=r_{\text{goal}}+r_{\text{grasp}}\), with both terms dynamically computed from simulator states to guide hand–object adjustments and optimize grasp execution.

To ensure consistency in data generation, we first assume that each object contains only a single type of joint, while allowing for multiple joints of that type. Furthermore, each joint type is associated with a predefined operational range of motion. The joint categories considered in our setting include the two most common types: revolute and prismatic joints.
For any grasp point \({p}\) relative to the object’s center of mass \(g_\mathcal{O}\), the grasp direction is \({D}={p}-{g}_\mathcal{O}\). Each frame provides a homogeneous transformation \(\mathcal{T}\in\mathbb{R}^{4\times4}\); we normalize coordinates by applying \(\mathcal{T}^{-1}\mathbf{x}\), mapping all grasping data into a unified canonical space.

To capture human intent, we render \textit{GraspXL} grasping sequences in \textit{Open3D} and prompt \textit{GPT-4} to produce multi-perspective textual descriptions covering contact points, wrist orientation, and intended action. This ensures concise, consistent annotations (Figure~\ref{fig:graspxlgen}). Our dataset comprises seven object categories, 256 object instances, and over 50,000 manipulation sequences in total.

\section{Experiments}
\subsection{Experimental Settings}
\noindent{\textbf{Implementation and metrics}}
We employ four codebooks to discretize hand and object representations. 
The manipulation language model is built upon Vicuna-7B-v1.5 and fine-tuned using LoRA. To comprehensively evaluate the quality of the generated grasping instructions from multiple perspectives, we adopt the evaluation methodology proposed by \citet{jiang2023motiongpt,huang2025hoigpt}. We consider the following evaluation metrics: {Fr\'{e}chet Inception Distance (FID), Diversity, Multi-modal Distance (MMDist), Interaction Volume (IV), Average Displacement Error (ADE), Final Displacement Error (FDE), Codebook Update Coverage (CUC). For detailed configurations and definitions, please refer to the supplementary material.


\begin{figure*}[htb]
  \centering
  \includegraphics[width=0.9\linewidth]{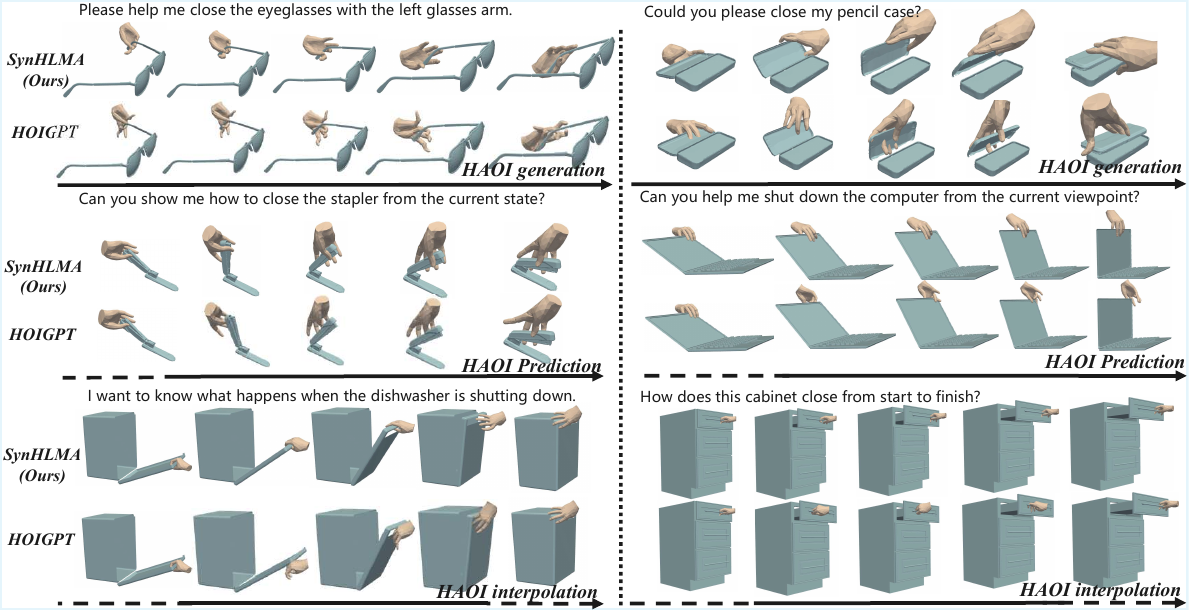}
  \caption{Qualitative results comparison between our SynHLMA and HOIGPT on three HAOI tasks of HAOI generation, HAOI prediction, and HAOI interpolation.}
  \label{fig:exp}
\end{figure*}

\begin{table*}[t]
\caption{Comparison with the state-of-the-art on HAOI generation. The $\rightarrow$ indicates closer to real is better, and $\downarrow$ indicates lower is better.}
\label{table:mainlab}
\setlength{\tabcolsep}{5pt}
\centering
\small 
\begin{tabular}{lcccccc}
\toprule
\bf Methods &\bf FID $\downarrow$ &\bf Diversity $\rightarrow$ &\bf MMDist $\downarrow$ &\bf IV $\downarrow$ &\bf ADE $\downarrow$ &\bf FDE $\downarrow$ 
\\ \midrule 
Real & 0.002 & 43.427 & - & - & - & - \\
T2MGPT~\cite{zhang2023generating} & 27.027 & 17.124 & 21.007 & 11.218 & 2.145 & 7.013 \\
MotionGPT~\cite{jiang2023motiongpt} & 27.361 & 16.876 & 21.625 & 11.436 & 2.119 & 4.879 \\
TM2T~\cite{guo2022tm2t} & 29.459 & 16.968 & 20.964 & 11.975 & 2.014 & 4.856 \\
Text2HOI~\cite{cha2024text2hoi} & 22.746 & 21.035 & 23.248 & 13.651 & 1.634 & 2.955 \\
SemGrasp~\cite{li2024semgrasp} & 20.873 & 27.954 & 16.482 & 10.247 & 1.284 & 1.932 \\
NL2Contact~\cite{zhang2024nl2contact} & 25.610 & 19.843 & 19.127 & 12.984 & 1.742 & 3.865 \\
HOIGPT~\cite{huang2025hoigpt} & 19.040 & 26.498 & 15.003 & 13.828 & 1.055 & 1.168 \\
SynHLMA & \textbf{14.121} & \textbf{40.484} & \textbf{12.793} & \textbf{5.919} & \textbf{0.976} & \textbf{1.147} \\
\bottomrule
\end{tabular}
\end{table*}

\begin{table*}[t]
\caption{Comparison on HAOI prediction and HAOI interpolation.}
\label{table:prediction}
\centering
\setlength{\tabcolsep}{3pt}    
\small 
\begin{tabular}{lcccc|ccc}
\toprule
\bf Methods &\bf FID $\downarrow$ &\bf Diversity $\rightarrow$ &\bf ADE $\downarrow$ &\bf FDE $\downarrow$ &\bf FID $\downarrow$ &\bf Diversity $\rightarrow$ &\bf ADE $\downarrow$
\\ \midrule 
Real & 0.001 & 48.283 & - & - & 0.001 & 45.465 & - \\
MotionGPT~\cite{jiang2023motiongpt} & 45.223 & 21.348 & 1.876 & 3.363 & 46.082 & 20.315 & 1.718 \\
HOIGPT~\cite{huang2025hoigpt} & 36.379 & 29.119 & 1.127 & \textbf{1.115} & 34.956 & 24.052 & 1.055 \\
SynHLMA & \textbf{21.739} & \textbf{48.691} & \textbf{0.968} & 1.125 & \textbf{25.225} & \textbf{44.012} & \textbf{0.986} \\
\bottomrule
\end{tabular}
\vspace{-4mm}
\end{table*}

\noindent{\textbf{Baselines}} HAOI generation refers to predicting the subsequent grasping sequence given the object’s point cloud and a corresponding textual description as input. HAOI prediction refers to the task where only the first 20\% of a grasping sequence is provided to the model, which must then predict the remaining 80\%. HAOI interpolation denotes the scenario in which 40–50\% of a sequence is intentionally omitted, requiring the model to complete the missing portions in a coherent manner. We compare SynHLMA with state-of-the-art HAOI generation models, including HOIGPT, Text2HOI, SemGrasp and NL2Contact, as well as with representative human motion generation baselines such as T2MGPT, MotionGPT, and TM2M.

\subsection{Comparison with State-of-the-arts} 
As shown in Table \ref{table:mainlab}, our method achieves state-of-the-art performance on the HAOI generation task, outperforming all baselines. Notably, it improves the FID score by \textbf{4.919\%} and achieves \textbf{12.530\%} increase in diversity.
Table \ref{table:prediction} presents results on the HAOI prediction and interpolation tasks. For HAOI prediction, our method achieves a \textbf{14.64\% }improvement in FID and outperforms baselines by \textbf{19.572\%} in diversity. In the interpolation task, we observe a \textbf{9.731\%} reduction in FID, along with a \textbf{19.969\%} gain in diversity.
Unlike prior models, our method introduces an articulation-guided decoder to explicitly capture articulated object structure, and employs a hierarchical HAOI token representation for fine-grained manipulation modeling. Finally, the proposed articulation-aware objective guides the model to generate grasps that are more consistent with the articulated object's physical state and kinematic configuration. Qualitative results of the proposed method are illustrated in Figure~\ref{fig:exp}.


\subsection{Ablation Studies}
\textbf{Articulation-Aware Objective Ablations} 
Table~\ref{table:articulationloss} reports the quantitative results of the articulation-aware objective. 
Removing $\mathcal{L}_{geom}$ increases FID to \textbf{15.872} and IV to \textbf{6.452}, while removing $\mathcal{L}_{temp}$ raises ADE to \textbf{1.112}; ablating both leads to the worst performance with FID \textbf{16.840}, IV \textbf{6.987}, and ADE \textbf{1.233}. 
These results demonstrate that our articulation-aware objective effectively enforces geometric validity and temporal consistency, enabling joint-state-aligned motion generation that faithfully follows articulated object dynamics.

\textbf{Multi-Level VQ-VAE Ablations} 
Table~\ref{table:vqvae} presents the ablation results on the multi-level VQ-VAE design. 
Increasing the codebook size to $\mathcal{K}=2048$ achieves the best FID of \textbf{0.913} and Diversity of \textbf{3.063}, while setting $d_{\mathcal{B}}=1024$ yields the lowest IV of \textbf{3.195}; even a compact codebook with $\mathcal{K}=512$ attains a CUC of \textbf{0.232}. 
This indicates that our hierarchical discrete representation preserves rich motion semantics while maintaining compactness, validating the effectiveness of structured codebook design for articulated grasp generation.

\noindent{\textbf{Token Semantics in Discrete Representations Ablations}} 
Table~\ref{table:token} summarizes the ablation results on token semantics. 
Our full design $\langle g,l,r,j\rangle$ achieves the best FID of \textbf{0.699} and ADE of \textbf{0.815}, outperforming variants such as removing all tokens or discarding semantic meanings. 
These findings confirm that explicitly encoding stage-wise semantic roles enables structured motion decomposition and enhances the model’s capacity to capture fine-grained grasping behaviors.

\textbf{Manipulation Language Model Ablations} Table~\ref{table:manipulation} reports the ablation results on the manipulation language model. 
Among different backbones, Gemma achieves FID \textbf{22.576} and IV \textbf{2.317}, while Llama yields FID \textbf{51.911}; inappropriate LoRA ranks severely degrade performance to FID \textbf{126.954}, and removing the two-stage training results in FID \textbf{39.849}. 
These results indicate that a properly configured backbone together with staged optimization stabilizes language-to-motion alignment, facilitating accurate mapping from textual manipulation intent to physically grounded action sequences.

\begin{table}[t]
\centering
\small
\caption{Ablation on articulation-aware objective.}
\label{table:articulationloss}
\setlength{\tabcolsep}{2.5pt} 
\begin{tabular}{lccc}
\toprule
& \bf FID $\downarrow$ & \bf IV $\downarrow$ & \bf ADE $\downarrow$ \\
\midrule 
w/o $\mathcal{L}_{geom}$ & 15.872 & 6.452 & 1.041 \\
w/o $\mathcal{L}_{temp}$ & 14.990 & 6.104 & 1.112 \\
w/o $\mathcal{L}_{geom} \land \mathcal{L}_{temp}$  & 16.840 & 6.987 & 1.233  \\
\bottomrule
\end{tabular}
\vspace{-4mm}
\end{table}

\begin{table}[t]
\centering
\small
\caption{Ablation on VQ-VAE design.}
\label{table:vqvae}
\setlength{\tabcolsep}{2pt} 
\begin{tabular}{lcccc}
\toprule
& \bf FID $\downarrow$ & \bf Diversity $\uparrow$ & \bf IV $\downarrow$ & \bf CUC \\
\midrule
$\mathcal{B}$ entries $\mathcal{K}$ = 512 & 1.237 & 2.944 & 3.690 & 0.232\\
$\mathcal{B}$ entries $\mathcal{K}$ = 2048 & \textbf{0.913} & \textbf{3.063} & 3.222 & 0.195\\
$\mathcal{B}$ dim. $d_\mathcal{B}$ = 256 & 1.113 & 3.704 & 3.446 & - \\
$\mathcal{B}$ dim. $d_\mathcal{B}$ = 1024 & 1.055 & 3.235 & \textbf{3.195} & - \\
EMA + Reset & 3.637 & 3.327 & 3.704 & -\\
\bottomrule
\end{tabular}
\vspace{-4mm}
\end{table}

\begin{table}[t]
\centering
\small
\caption{Ablation on discrete representation.}
\label{table:token}
\setlength{\tabcolsep}{3pt} 
\begin{tabular}{lccc}
\toprule
& \bf FID $\downarrow$ & \bf Diversity $\uparrow$ & \bf ADE $\downarrow$ \\
\midrule 
$\langle g,l \rangle$ & 0.976 & \textbf{3.428} & 0.974  \\
$\langle g,l,r \rangle$ & 1.152 & 3.379 & 0.988 \\
$\langle g,l,r \times 2,j \rangle$ & 1.696 & 2.790 & 0.989  \\
$\langle g,l,r \times 3,j \rangle$ & 1.179 & 2.715 & 0.893 \\
w/o $\langle g,l,r,j \rangle$ & 1.160 & 3.411 & 1.012  \\
w/o semantic & 1.055 & 3.244 & 0.948  \\
Shared Codebook & 0.940 & 3.086 & 0.925 \\
$\langle g,l,r,j\rangle$ (Ours) & \textbf{0.699} & 3.109 & \textbf{0.815} \\
\bottomrule
\end{tabular}
\vspace{-4mm}
\end{table}

\begin{table}[t]
\centering
\small
\caption{Ablation on manipulation language model setting.}
\label{table:manipulation}
\setlength{\tabcolsep}{3pt} 
\begin{tabular}{lccc}
\toprule
 & \bf FID $\downarrow$ & \bf MMDist $\downarrow$ & \bf IV $\downarrow$  \\
\midrule
w/ Llama & 51.911 & 17.222 & 9.244  \\
w/ Qwen & 73.590 & 25.272 & 6.848  \\
w/ Gemma & 22.576 & 16.228 & 2.317  \\
LoRA $r=8$ & 126.954 & 21.020 & 7.703 \\
LoRA $r=32$ & 53.485 & 18.737 & 7.580 \\
w/o 2-stage & 39.849 & 19.472 & 7.237 \\
\bottomrule
\end{tabular}
\vspace{-4mm}
\end{table}

\begin{figure}[htb]
  \centering
  \includegraphics[width=0.9\linewidth]{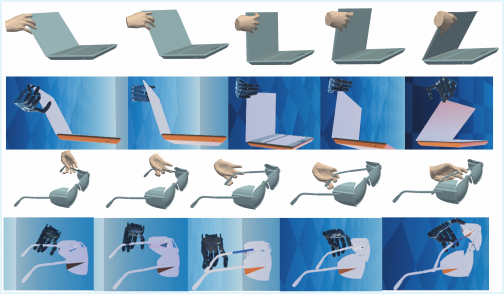}
  \caption{Application of our SynHLMA transferring into ShadowHand model in the robotics scenario}
  \label{fig:raisim}
  \vspace{-4mm}
\end{figure}
 
\subsection{Dexterous Manipulation Transfer to Robotics}
We demonstrate the application of our SynHLMA in embodied robotics area, and transfer the learned manipulation sequence poses into ShadowHand model~\cite{li2019vision} within RaiSim simulator in Fig. \ref{fig:raisim}. In detail, we align the finger keypoints from MANO model generated by SynHLMA and ShadowHand model with a simple fitting-based optimization approach. Next, using the fitted ShadowHand poses, we can enable the dexterous manipulation that implements on the same articulated objects. 
\vspace{-4mm}
\section{Conclusion}
In this work, we propose \textbf{SynHLMA}, a novel manipulation language model for articulated objects based on discrete representations. Moreover, an \textbf{articulation-aware objective} is incorporated to enforce geometric and physical validity, achieve representation-level joint-state alignment, and enhance the model’s responsiveness to articulated object variations. To support this framework, we introduce \textbf{HAOI-Lang}, a new dataset that bridges the gap between manipulation actions and their corresponding language instructions. By designing multi-level tokens, our model captures fine-grained grasping processes, achieving strong performance across various downstream tasks. Furthermore, we demonstrate its potential in enhancing dexterous grasping for robotic hands. In future work, we aim to explore more fine-grained and coordinated bimanual manipulation.

\section*{Impact Statement}

This paper presents work whose goal is to advance the field of Machine Learning. There are many potential societal consequences of our work, none which we feel must be specifically highlighted here.

\nocite{langley00}

\bibliography{example_paper}

@article{jin2024reasoning,
  title={Reasoning grasping via multimodal large language model},
  author={Jin, Shiyu and Xu, Jinxuan and Lei, Yutian and Zhang, Liangjun},
  journal={arXiv preprint arXiv:2402.06798},
  year={2024}
}

@inproceedings{yang2022artiboost,
  title={Artiboost: Boosting articulated 3d hand-object pose estimation via online exploration and synthesis},
  author={Yang, Lixin and Li, Kailin and Zhan, Xinyu and Lv, Jun and Xu, Wenqiang and Li, Jiefeng and Lu, Cewu},
  booktitle={Proceedings of the IEEE/CVF conference on computer vision and pattern recognition},
  pages={2750--2760},
  year={2022}
}

@article{team2024gemma,
  title={Gemma 2: Improving open language models at a practical size},
  author={Team, Gemma and Riviere, Morgane and Pathak, Shreya and Sessa, Pier Giuseppe and Hardin, Cassidy and Bhupatiraju, Surya and Hussenot, L{\'e}onard and Mesnard, Thomas and Shahriari, Bobak and Ram{\'e}, Alexandre and others},
  journal={arXiv preprint arXiv:2408.00118},
  year={2024}
}

@inproceedings{karunratanakul2021skeleton,
  title={A skeleton-driven neural occupancy representation for articulated hands},
  author={Karunratanakul, Korrawe and Spurr, Adrian and Fan, Zicong and Hilliges, Otmar and Tang, Siyu},
  booktitle={2021 International Conference on 3D Vision (3DV)},
  pages={11--21},
  year={2021},
  organization={IEEE}
}

@article{jiang2023motiongpt,
  title={Motiongpt: Human motion as a foreign language},
  author={Jiang, Biao and Chen, Xin and Liu, Wen and Yu, Jingyi and Yu, Gang and Chen, Tao},
  journal={Advances in Neural Information Processing Systems},
  volume={36},
  pages={20067--20079},
  year={2023}
}

@inproceedings{mo2019partnet,
  title={Partnet: A large-scale benchmark for fine-grained and hierarchical part-level 3d object understanding},
  author={Mo, Kaichun and Zhu, Shilin and Chang, Angel X and Yi, Li and Tripathi, Subarna and Guibas, Leonidas J and Su, Hao},
  booktitle={Proceedings of the IEEE/CVF conference on computer vision and pattern recognition},
  pages={909--918},
  year={2019}
}

@inproceedings{devlin2019bert,
  title={Bert: Pre-training of deep bidirectional transformers for language understanding},
  author={Devlin, Jacob and Chang, Ming-Wei and Lee, Kenton and Toutanova, Kristina},
  booktitle={Proceedings of the 2019 conference of the North American chapter of the association for computational linguistics: human language technologies, volume 1 (long and short papers)},
  pages={4171--4186},
  year={2019}
}

@article{achiam2023gpt,
  title={Gpt-4 technical report},
  author={Achiam, Josh and Adler, Steven and Agarwal, Sandhini and Ahmad, Lama and Akkaya, Ilge and Aleman, Florencia Leoni and Almeida, Diogo and Altenschmidt, Janko and Altman, Sam and Anadkat, Shyamal and others},
  journal={arXiv preprint arXiv:2303.08774},
  year={2023}
}

@article{bai2023qwen,
  title={Qwen technical report},
  author={Bai, Jinze and Bai, Shuai and Chu, Yunfei and Cui, Zeyu and Dang, Kai and Deng, Xiaodong and Fan, Yang and Ge, Wenbin and Han, Yu and Huang, Fei and others},
  journal={arXiv preprint arXiv:2309.16609},
  year={2023}
}

@article{team2023gemini,
  title={Gemini: a family of highly capable multimodal models},
  author={Team, Gemini and Anil, Rohan and Borgeaud, Sebastian and Alayrac, Jean-Baptiste and Yu, Jiahui and Soricut, Radu and Schalkwyk, Johan and Dai, Andrew M and Hauth, Anja and Millican, Katie and others},
  journal={arXiv preprint arXiv:2312.11805},
  year={2023}
}

@inproceedings{zhang2020pegasus,
  title={Pegasus: Pre-training with extracted gap-sentences for abstractive summarization},
  author={Zhang, Jingqing and Zhao, Yao and Saleh, Mohammad and Liu, Peter},
  booktitle={International conference on machine learning},
  pages={11328--11339},
  year={2020},
  organization={PMLR}
}

@article{borsos2023audiolm,
  title={Audiolm: a language modeling approach to audio generation},
  author={Borsos, Zal{\'a}n and Marinier, Rapha{\"e}l and Vincent, Damien and Kharitonov, Eugene and Pietquin, Olivier and Sharifi, Matt and Roblek, Dominik and Teboul, Olivier and Grangier, David and Tagliasacchi, Marco and others},
  journal={IEEE/ACM transactions on audio, speech, and language processing},
  volume={31},
  pages={2523--2533},
  year={2023},
  publisher={IEEE}
}

@article{li2023videochat,
  title={Videochat: Chat-centric video understanding},
  author={Li, KunChang and He, Yinan and Wang, Yi and Li, Yizhuo and Wang, Wenhai and Luo, Ping and Wang, Yali and Wang, Limin and Qiao, Yu},
  journal={arXiv preprint arXiv:2305.06355},
  year={2023}
}

@inproceedings{du2022glam,
  title={Glam: Efficient scaling of language models with mixture-of-experts},
  author={Du, Nan and Huang, Yanping and Dai, Andrew M and Tong, Simon and Lepikhin, Dmitry and Xu, Yuanzhong and Krikun, Maxim and Zhou, Yanqi and Yu, Adams Wei and Firat, Orhan and others},
  booktitle={International conference on machine learning},
  pages={5547--5569},
  year={2022},
  organization={PMLR}
}

@inproceedings{zhang2024nl2contact,
  title={NL2Contact: Natural Language Guided 3D Hand-Object Contact Modeling with Diffusion Model},
  author={Zhang, Zhongqun and Wang, Hengfei and Yu, Ziwei and Cheng, Yihua and Yao, Angela and Chang, Hyung Jin},
  booktitle={European Conference on Computer Vision},
  pages={284--300},
  year={2024},
  organization={Springer}
}

@inproceedings{huang2025hoigpt,
  title={HOIGPT: Learning Long-Sequence Hand-Object Interaction with Language Models},
  author={Huang, Mingzhen and Chu, Fu-Jen and Tekin, Bugra and Liang, Kevin J and Ma, Haoyu and Wang, Weiyao and Chen, Xingyu and Gleize, Pierre and Xue, Hongfei and Lyu, Siwei and others},
  booktitle={Proceedings of the Computer Vision and Pattern Recognition Conference},
  pages={7136--7146},
  year={2025}
}

@inproceedings{yang2021cpf,
  title={Cpf: Learning a contact potential field to model the hand-object interaction},
  author={Yang, Lixin and Zhan, Xinyu and Li, Kailin and Xu, Wenqiang and Li, Jiefeng and Lu, Cewu},
  booktitle={Proceedings of the IEEE/CVF International Conference on Computer Vision},
  pages={11097--11106},
  year={2021}
}

@inproceedings{prokudin2019efficient,
  title={Efficient learning on point clouds with basis point sets},
  author={Prokudin, Sergey and Lassner, Christoph and Romero, Javier},
  booktitle={Proceedings of the IEEE/CVF international conference on computer vision},
  pages={4332--4341},
  year={2019}
}

@inproceedings{liu2022akb,
  title={Akb-48: A real-world articulated object knowledge base},
  author={Liu, Liu and Xu, Wenqiang and Fu, Haoyuan and Qian, Sucheng and Yu, Qiaojun and Han, Yang and Lu, Cewu},
  booktitle={Proceedings of the IEEE/CVF Conference on Computer Vision and Pattern Recognition},
  pages={14809--14818},
  year={2022}
}

@inproceedings{zhang2024graspxl,
  title={Graspxl: Generating grasping motions for diverse objects at scale},
  author={Zhang, Hui and Christen, Sammy and Fan, Zicong and Hilliges, Otmar and Song, Jie},
  booktitle={European Conference on Computer Vision},
  pages={386--403},
  year={2024},
  organization={Springer}
}

@inproceedings{christen2022d,
  title={D-grasp: Physically plausible dynamic grasp synthesis for hand-object interactions},
  author={Christen, Sammy and Kocabas, Muhammed and Aksan, Emre and Hwangbo, Jemin and Song, Jie and Hilliges, Otmar},
  booktitle={Proceedings of the IEEE/CVF Conference on Computer Vision and Pattern Recognition},
  pages={20577--20586},
  year={2022}
}

@inproceedings{ye2023affordance,
  title={Affordance diffusion: Synthesizing hand-object interactions},
  author={Ye, Yufei and Li, Xueting and Gupta, Abhinav and De Mello, Shalini and Birchfield, Stan and Song, Jiaming and Tulsiani, Shubham and Liu, Sifei},
  booktitle={Proceedings of the IEEE/CVF Conference on Computer Vision and Pattern Recognition},
  pages={22479--22489},
  year={2023}
}

@article{deitke2023objaverse,
  title={Objaverse-xl: A universe of 10m+ 3d objects},
  author={Deitke, Matt and Liu, Ruoshi and Wallingford, Matthew and Ngo, Huong and Michel, Oscar and Kusupati, Aditya and Fan, Alan and Laforte, Christian and Voleti, Vikram and Gadre, Samir Yitzhak and others},
  journal={Advances in Neural Information Processing Systems},
  volume={36},
  pages={35799--35813},
  year={2023}
}

@inproceedings{xiang2020sapien,
  title={Sapien: A simulated part-based interactive environment},
  author={Xiang, Fanbo and Qin, Yuzhe and Mo, Kaichun and Xia, Yikuan and Zhu, Hao and Liu, Fangchen and Liu, Minghua and Jiang, Hanxiao and Yuan, Yifu and Wang, He and others},
  booktitle={Proceedings of the IEEE/CVF conference on computer vision and pattern recognition},
  pages={11097--11107},
  year={2020}
}

@article{hwangbo2018per,
  title={Per-contact iteration method for solving contact dynamics},
  author={Hwangbo, Jemin and Lee, Joonho and Hutter, Marco},
  journal={IEEE Robotics and Automation Letters},
  volume={3},
  number={2},
  pages={895--902},
  year={2018},
  publisher={IEEE}
}

@article{hussain2020unity,
  title={Unity game development engine: A technical survey},
  author={Hussain, Afzal and Shakeel, Haad and Hussain, Faizan and Uddin, Nasir and Ghouri, Turab Latif},
  journal={Univ. Sindh J. Inf. Commun. Technol},
  volume={4},
  number={2},
  pages={73--81},
  year={2020}
}

@inproceedings{radford2021learning,
  title={Learning transferable visual models from natural language supervision},
  author={Radford, Alec and Kim, Jong Wook and Hallacy, Chris and Ramesh, Aditya and Goh, Gabriel and Agarwal, Sandhini and Sastry, Girish and Askell, Amanda and Mishkin, Pamela and Clark, Jack and others},
  booktitle={International conference on machine learning},
  pages={8748--8763},
  year={2021},
  organization={PmLR}
}

@article{feix2015grasp,
  title={The grasp taxonomy of human grasp types},
  author={Feix, Thomas and Romero, Javier and Schmiedmayer, Heinz-Bodo and Dollar, Aaron M and Kragic, Danica},
  journal={IEEE Transactions on human-machine systems},
  volume={46},
  number={1},
  pages={66--77},
  year={2015},
  publisher={IEEE}
}

@article{xie2023learning,
  title={Learning-based robotic grasping: A review},
  author={Xie, Zhen and Liang, Xinquan and Roberto, Canale},
  journal={Frontiers in Robotics and AI},
  volume={10},
  pages={1038658},
  year={2023},
  publisher={Frontiers Media SA}
}

@inproceedings{grady2021contactopt,
  title={Contactopt: Optimizing contact to improve grasps},
  author={Grady, Patrick and Tang, Chengcheng and Twigg, Christopher D and Vo, Minh and Brahmbhatt, Samarth and Kemp, Charles C},
  booktitle={Proceedings of the IEEE/CVF Conference on Computer Vision and Pattern Recognition},
  pages={1471--1481},
  year={2021}
}

@inproceedings{yang2022oakink,
  title={Oakink: A large-scale knowledge repository for understanding hand-object interaction},
  author={Yang, Lixin and Li, Kailin and Zhan, Xinyu and Wu, Fei and Xu, Anran and Liu, Liu and Lu, Cewu},
  booktitle={Proceedings of the IEEE/CVF conference on computer vision and pattern recognition},
  pages={20953--20962},
  year={2022}
}

@inproceedings{li2024semgrasp,
  title={Semgrasp: Semantic grasp generation via language aligned discretization},
  author={Li, Kailin and Wang, Jingbo and Yang, Lixin and Lu, Cewu and Dai, Bo},
  booktitle={European Conference on Computer Vision},
  pages={109--127},
  year={2024},
  organization={Springer}
}

@inproceedings{xu2023unidexgrasp,
  title={Unidexgrasp: Universal robotic dexterous grasping via learning diverse proposal generation and goal-conditioned policy},
  author={Xu, Yinzhen and Wan, Weikang and Zhang, Jialiang and Liu, Haoran and Shan, Zikang and Shen, Hao and Wang, Ruicheng and Geng, Haoran and Weng, Yijia and Chen, Jiayi and others},
  booktitle={Proceedings of the IEEE/CVF Conference on Computer Vision and Pattern Recognition},
  pages={4737--4746},
  year={2023}
}

@misc{xue2021omadobjectmodelarticulated,
      title={OMAD: Object Model with Articulated Deformations for Pose Estimation and Retrieval}, 
      author={Han Xue and Liu Liu and Wenqiang Xu and Haoyuan Fu and Cewu Lu},
      year={2021},
      eprint={2112.07334},
      archivePrefix={arXiv},
      primaryClass={cs.CV},
      url={https://arxiv.org/abs/2112.07334}, 
}

@inproceedings{zhang2023generating,
  title={Generating human motion from textual descriptions with discrete representations},
  author={Zhang, Jianrong and Zhang, Yangsong and Cun, Xiaodong and Zhang, Yong and Zhao, Hongwei and Lu, Hongtao and Shen, Xi and Shan, Ying},
  booktitle={Proceedings of the IEEE/CVF conference on computer vision and pattern recognition},
  pages={14730--14740},
  year={2023}
}

@inproceedings{guo2022tm2t,
  title={Tm2t: Stochastic and tokenized modeling for the reciprocal generation of 3d human motions and texts},
  author={Guo, Chuan and Zuo, Xinxin and Wang, Sen and Cheng, Li},
  booktitle={European Conference on Computer Vision},
  pages={580--597},
  year={2022},
  organization={Springer}
}

@inproceedings{cha2024text2hoi,
  title={Text2hoi: Text-guided 3d motion generation for hand-object interaction},
  author={Cha, Junuk and Kim, Jihyeon and Yoon, Jae Shin and Baek, Seungryul},
  booktitle={Proceedings of the IEEE/CVF Conference on Computer Vision and Pattern Recognition},
  pages={1577--1585},
  year={2024}
}

@inproceedings{li2019vision,
  title={Vision-based teleoperation of shadow dexterous hand using end-to-end deep neural network},
  author={Li, Shuang and Ma, Xiaojian and Liang, Hongzhuo and G{\"o}rner, Michael and Ruppel, Philipp and Fang, Bin and Sun, Fuchun and Zhang, Jianwei},
  booktitle={2019 International Conference on Robotics and Automation (ICRA)},
  pages={416--422},
  year={2019},
  organization={IEEE}
}
\bibliographystyle{icml2026}

\newpage
\appendix
\onecolumn
\section{Overview}
In this supplementary material, we first describe in detail the usage of Large Language Models (LLMs) in our paper. We then provide a comprehensive explanation of the composition and generation method of the HAOI-Lang dataset. Next, we present an in-depth description of the experimental setup, including batch loss functions and scoring metrics not covered in the main paper. Finally, we showcase the visualization results of our Discrete Articulated Manipulation Representation across multiple downstream HAOI tasks, demonstrating its generalization ability across different objects within the same category.


\section{Use of Large Language Models}

In the preparation of this paper, we used ChatGPT and DeepSeek to assist with language polishing. We also consulted ChatGPT for inspiration regarding the code architecture. All substantive research design, implementation, data analysis, and writing decisions were made by the authors. Every output generated with the help of LLMs was carefully reviewed, verified, and revised by the authors. The authors retain full responsibility for all content in this paper.

\section{HAOI-Lang Dataset Construction}
\subsection{Dataset Composition and Object Coverage}
Our HAOI-Lan dataset consists of seven commonly encountered articulated objects: \textbf{stapler}, \textbf{laptop}, \textbf{scissors}, \textbf{cabinet}, \textbf{dishwasher}, \textbf{eyeglasses}, and \textbf{box}. These objects collectively cover nearly all major types of articulated structures found in real-world environments, such as rotational joints, sliding mechanisms, and compound linkages. Each object category includes around 30–70 object instances, ensuring significant intra-class variation in appearance, articulation, and geometry.

For each object instance, we first segment the graspable regions from the non-graspable parts. Within these graspable regions, we \textbf{uniformly sample 200 surface points}. Around each sampled point, we simulate or generate a batch of manipulation sequences, capturing plausible interactions grounded in object geometry and affordance. To enhance semantic richness, we leverage GPT-4 to generate a unique caption for each manipulation episode, describing the intent or action in natural language. The detailed statistics of the dataset are shown in Table \ref{tab:haoi_stats}.

\begin{table}[h]
\caption{Statistics of the HAOI-Lang dataset. Each object category contains multiple instances, and each instance includes 200 manipulation episodes paired with 200 unique GPT-4-generated captions.}
\label{tab:haoi_stats}
\centering
\begin{tabular}{lcccc}
\toprule
\textbf{Category} & \textbf{Instances} & \textbf{Angle} & \textbf{Manipulations \& Captions} \\
\midrule
Stapler        & 8 & 10 & 1600  \\
Box            & 8 & 10 & 1600  \\
Laptop         & 55 & 10 & 11000  \\
Scissors       & 45 & 10 & 9000  \\
Cabinet      & 38 & 10 & 7600  \\
Dishwasher     & 37 & 10 & 7400  \\
Eyeglasses     & 65 & 10 & 13000  \\
\midrule
\textbf{Total} & \textbf{256} & - &\textbf{51200} \\
\bottomrule
\end{tabular}
\end{table}

\subsection{Semantic Captioning and Instructional Dialogue Generation}
As described in the main paper, we use Open3D to visualize each manipulation episode into a series of sequential image frames. These frame sequences are then passed to ChatGPT-4, which generates a detailed natural language description of the action, capturing the spatial relations, articulation dynamics, and hand-object interactions.

To align with the Instruction-to-Motion task format, we further utilize GPT-4 to transform the detailed captions into instructional dialogues—specifically, concise directives issued from an instructor to an executor. We designed a specialized prompt to guide GPT-4 in generating this dialogue format, with the goal of enriching our dataset with task-driven, goal-oriented language instructions.

An example of such a dialogue template is shown in Figure \ref{fig:gpt}. 

\begin{figure}[htb]
  \centering
  \includegraphics[width=1\linewidth]{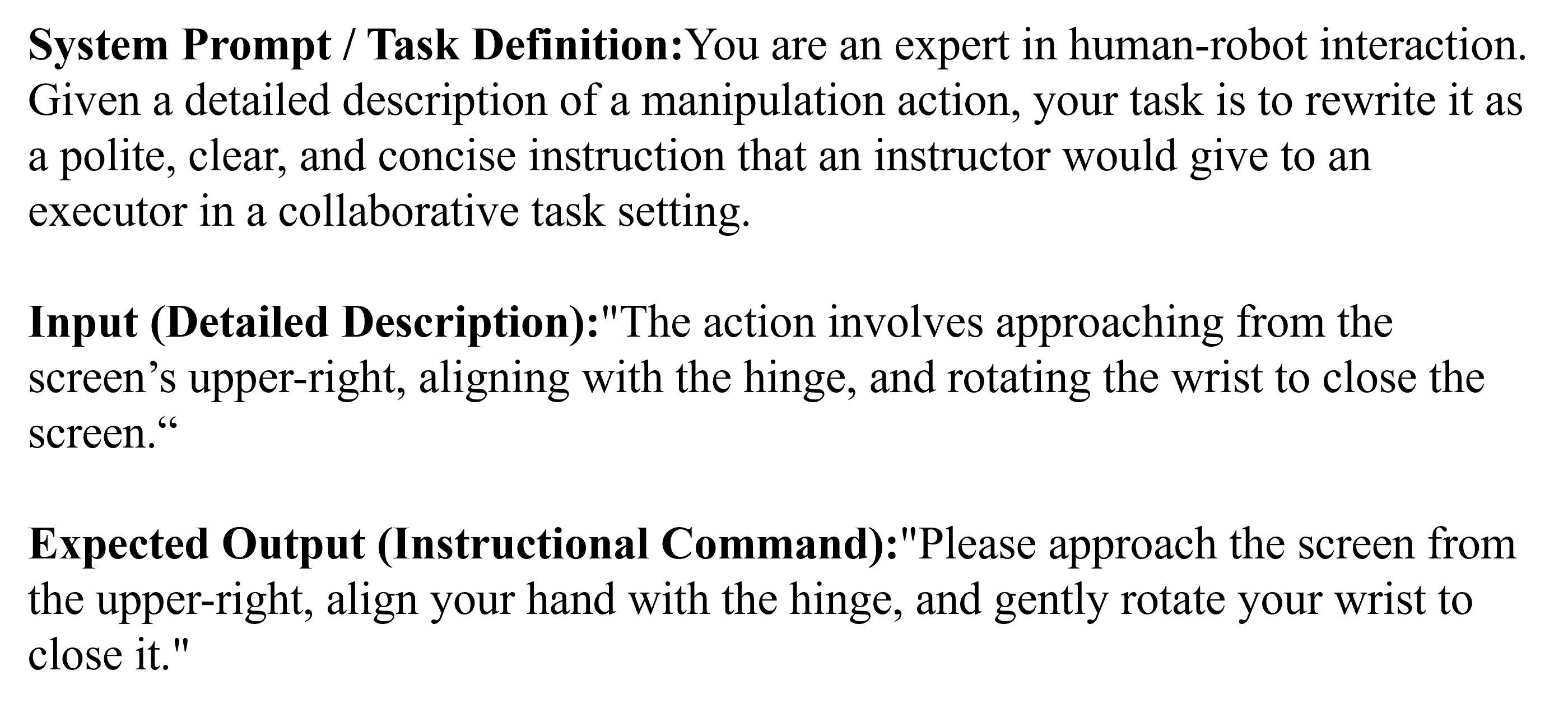}
  \caption{Instructional dialogue template created by GPT-4 for HAOI-Lang.}
  \label{fig:gpt}
\end{figure}

\section{Experiments Details}
\subsection{Implementation Details}
\textbf{Discrete Articulated Manipulation Representation} For this task, we employ a multi-level VQ-VAE where each codebook is of size 1024x512. There are four codebooks in total, each corresponding to different levels of articulation: global pose, local articulation, and refinement of both hand and object poses. The training configuration for this model includes:

\begin{itemize}
  \item \textbf{Learning Rate:} 0.00002
  \item \textbf{Batch Size:} 128
  \item \textbf{Number of Workers:} 8
  \item \textbf{Number of Points:} 2048
  \item \textbf{Pose Loss:} 1.0
  \item \textbf{$\lambda_\text{pen}$ :} 1.0
  \item \textbf{$\lambda_\text{joint}$ :} 1.0
  \item \textbf{$\lambda_\text{temp}$ :} 1.0
  \item \textbf{$\lambda_1$ :} 1.0
  \item \textbf{$\lambda_2$ :} 1.0
  \item \textbf{$\lambda_3$ :} 1.0
  \item \textbf{$\lambda_4$ :} 1.0

\end{itemize}

\textbf{HAOI Manipulation Language Model} To equip the model with the ability to understand and generate language instructions for manipulation tasks, we fine-tuned the Vicuna-7B v1.5 model using parameter-efficient adaptation via LoRA. The input to the model consists of manipulation instructions, and the output is a structured semantic representation related to object interaction. The model was trained on high-quality instruction-action paired data with the following key settings:
\begin{itemize}
  \item \textbf{Base Model:} Vicuna-7B
  \item \textbf{Epochs:} 20
  \item \textbf{Batch Size:} 128 (with a micro batch size of 6, using gradient accumulation)
  \item \textbf{Learning Rate:} 3e-4
  \item \textbf{Maximum Input Length:} 256 tokens
  \item \textbf{Validation Set Size:} 11,000
  \item \textbf{LoRA Rank ($r$):} 16
  \item \textbf{Scaling Factor ($\alpha$):} 32
  \item \textbf{Dropout:} 0.05
  \item \textbf{Target Modules:} \texttt{q\_proj}, \texttt{v\_proj}, \texttt{embed\_tokens}, \texttt{lm\_head}
\end{itemize}

\subsection{Evaluation Metrics}

To complement the evaluation methodology discussed in the main paper, we provide a detailed account of the quantitative metrics used to assess the quality, diversity, and accuracy of the generated hand-object interaction sequences. These metrics are inspired by previous work in instruction-conditioned motion generation~\cite{jiang2023motiongpt,huang2025hoigpt} and are tailored to capture both spatial precision and behavioral richness. In addition, to verify that the capacity of the VQ-VAE can adequately discretize continuous tasks, we designed a metric to measure its utilization rate.

\textbf{Fr\'{e}chet Inception Distance (FID)} is used to measure the distributional similarity between the real and generated interaction trajectories. Specifically, it compares the feature embeddings (typically from a pretrained 3D encoder or motion encoder) of both sets using the Fréchet distance:

\begin{equation}
\text{FID} = \|\mu_r - \mu_g\|_2^2 + \text{Tr}(\Sigma_r + \Sigma_g - 2(\Sigma_r \Sigma_g)^{1/2})
\end{equation}

where $(\mu_r, \Sigma_r)$ and $(\mu_g, \Sigma_g)$ are the means and covariances of real and generated features, respectively. Lower FID scores indicate higher generation fidelity.

\textbf{Diversity} reflects how varied the generated motion trajectories are across different random seeds or conditions. It is calculated as the average pairwise cosine distance between the generated feature vectors:

\begin{equation}
\text{Diversity} = \frac{2}{N(N-1)} \sum_{i < j} \left(1 - \frac{f_i \cdot f_j}{\|f_i\| \|f_j\|} \right)
\end{equation}

Higher diversity scores indicate a richer variety of generated behaviors.

\textbf{MultiModality (MModality)} evaluates the variability of outputs when the input instruction is held constant. It is computed as the average pairwise distance between generated samples conditioned on the same instruction:

\begin{equation}
\text{MModality} = \frac{2}{K(K-1)} \sum_{i < j} \|x_i - x_j\|_2
\end{equation}

where $x_i$ and $x_j$ are generated trajectories under the same condition, and $K$ is the number of generated samples per condition.

\textbf{Interaction Volume (IV)} measures how much of the object’s surface is covered by the hand during interaction. It quantifies the cumulative 3D space traversed by the hand that is in proximity to the object, typically computed via voxelization:

\begin{equation}
\text{IV} = \sum_{v \in \mathcal{V}} \mathbf{1}\left[\text{dist}(v, H) < \epsilon \right]
\end{equation}

where $\mathcal{V}$ is the set of object voxels, $H$ is the hand trajectory, and $\epsilon$ is a distance threshold (e.g., $1\,\mathrm{cm}$). The result is reported in $\mathrm{cm}^3$.

\textbf{Average Displacement Error (ADE)} quantifies the temporal accuracy of the predicted hand trajectory with respect to the ground truth. It is defined as the mean Euclidean distance across all time steps:

\begin{equation}
\text{ADE} = \frac{1}{T} \sum_{t=1}^{T} \| x_t - \hat{x}_t \|_2
\end{equation}

where $x_t$ and $\hat{x}_t$ denote the predicted and ground-truth hand positions at time step $t$.

\textbf{Final Displacement Error (FDE)} captures the spatial deviation at the final frame of the trajectory:

\begin{equation}
\text{FDE} = \| x_T - \hat{x}_T \|_2
\end{equation}

FDE is especially important for evaluating whether the generated trajectory reaches the intended end state.

\textbf{Codebook Update Coverage (CUC)} is used to measure the proportion of codebook entries that are actively updated during each training or evaluation round. Specifically, it quantifies how extensively the discrete latent space of a VQ-VAE is utilized:

\begin{equation}
\text{CUC} = \frac{N_\text{updated}}{N_\text{total}}
\end{equation}

where \(N_\text{updated}\) is the number of codebook entries updated in the current round and \(N_\text{total}\) is the total number of entries in the codebook. Higher CUC values indicate broader utilization of the codebook and thus better coverage of the latent space.

\section{Generalization Across Object Scales in HAOI Tasks}

To demonstrate the diversity and generalization ability of our generative model, we visualize three types of downstream tasks—\textit{HAOI generation}, \textit{HAOI prediction}, and \textit{HAOI interpolation}—on four representative articulated objects. These include two small- to medium-sized objects: \textbf{glasses} and a \textbf{laptop}, and two large-scale objects: a \textbf{cabinet} and a \textbf{dishwasher}.

The visualizations (see Figure \ref{fig:small_objects_1}, Figure \ref{fig:small_objects_2}, Figure \ref{fig:large_objects_1} and \ref{fig:large_objects_2}) show that:
\begin{itemize}
  \item Our model is capable of generating \textbf{distinct and plausible HAOI motions at different locations} on the same object, effectively adapting to local geometry.
  \item Even when the \textbf{grasping location is fixed}, the model can produce \textbf{diverse grasp directions}, reflecting its ability to represent multimodal interaction possibilities.
  \item Importantly, the model maintains \textbf{robust performance even on large articulated objects}, successfully generating and inferring physically plausible hand-object interactions without degradation in quality.
\end{itemize}

These results highlight the flexibility of our discrete representation and its effectiveness in modeling a wide range of manipulation behaviors across object scales and task types.

\section{Failure Cases in HAOI Manipulation}
We select the most representative failure cases from the test set, as illustrated in Fig.~\ref{fig:failurecase}. 
In \textbf{HAOI generation}, when the manipulation description is overly strict or physically implausible, the resulting grasp may deviate from realistic or physically feasible configurations. 
In \textbf{HAOI prediction}, the predicted interaction sequence may exhibit temporal inconsistencies, such as hand drift from the intended grasping position or deformation of the grasp over time. 
In \textbf{HAOI interpolation}, the primary issue is abrupt changes in hand pose, which are often caused by discontinuities or invalid predictions in the interpolated pose sequence.

\begin{figure*}[htb]
  \centering
  \includegraphics[width=1\linewidth]{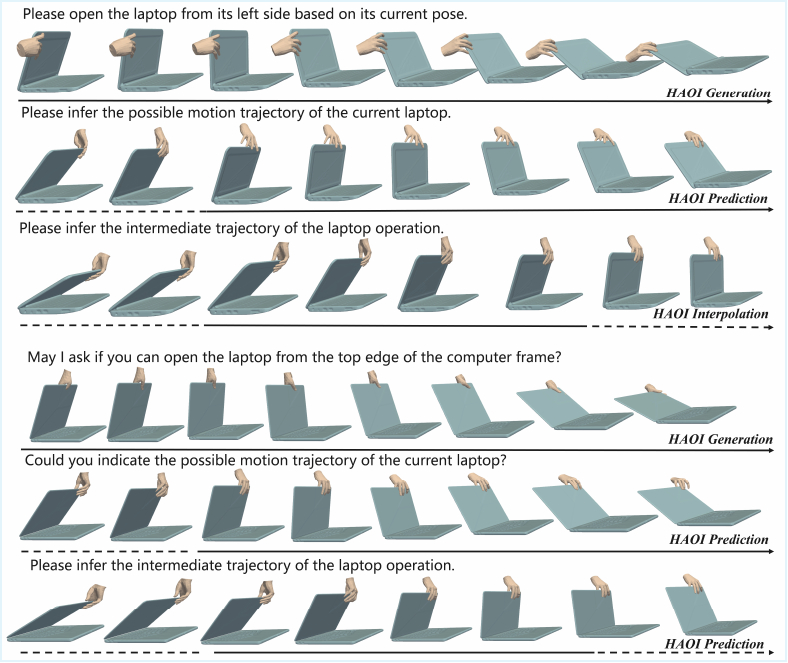}
  \caption{HAOI generation, prediction, and completion on small-to-medium objects: laptop.} 
  \label{fig:small_objects_1}
\end{figure*}
\begin{figure*}[htb]
  \centering
  \includegraphics[width=1\linewidth]{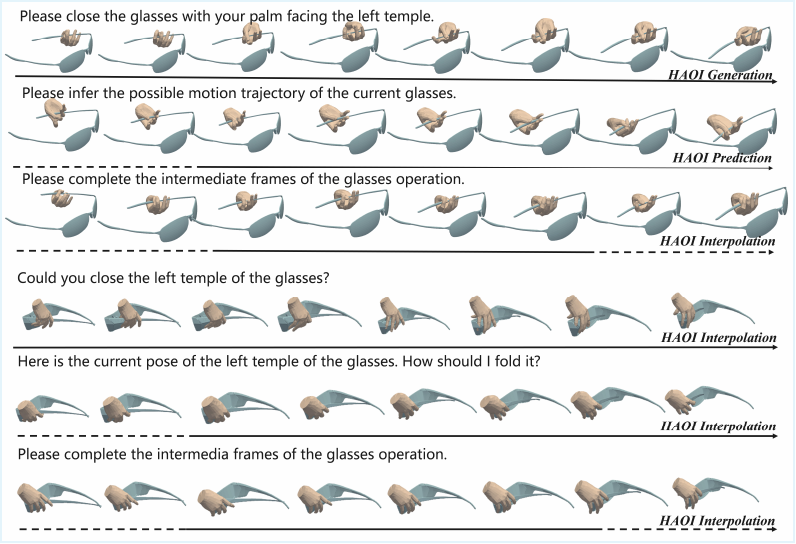}
  \caption{HAOI generation, prediction, and completion on small-to-medium objects: glasses.} 
  \label{fig:small_objects_2}
\end{figure*}

\begin{figure*}[htb]
  \centering
  \includegraphics[width=1\linewidth]{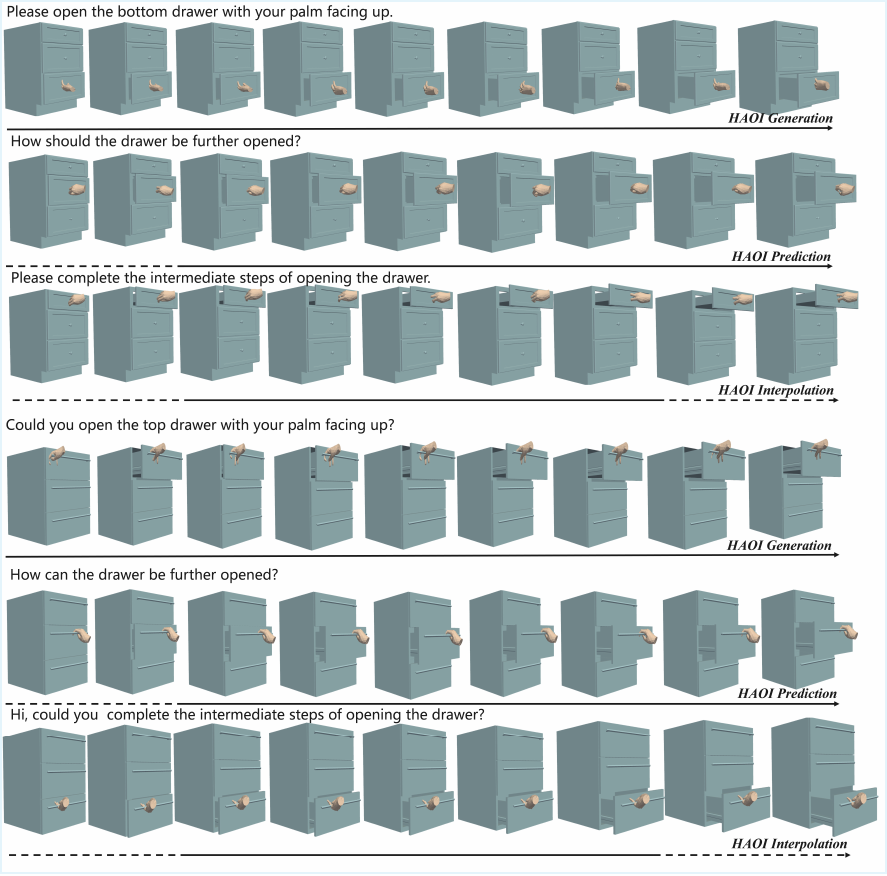}
  \caption{HAOI generation, prediction, and completion on large objects: cabinet.}
  \label{fig:large_objects_1}
\end{figure*}

\begin{figure*}[htb]
  \centering
  \includegraphics[width=1\linewidth]{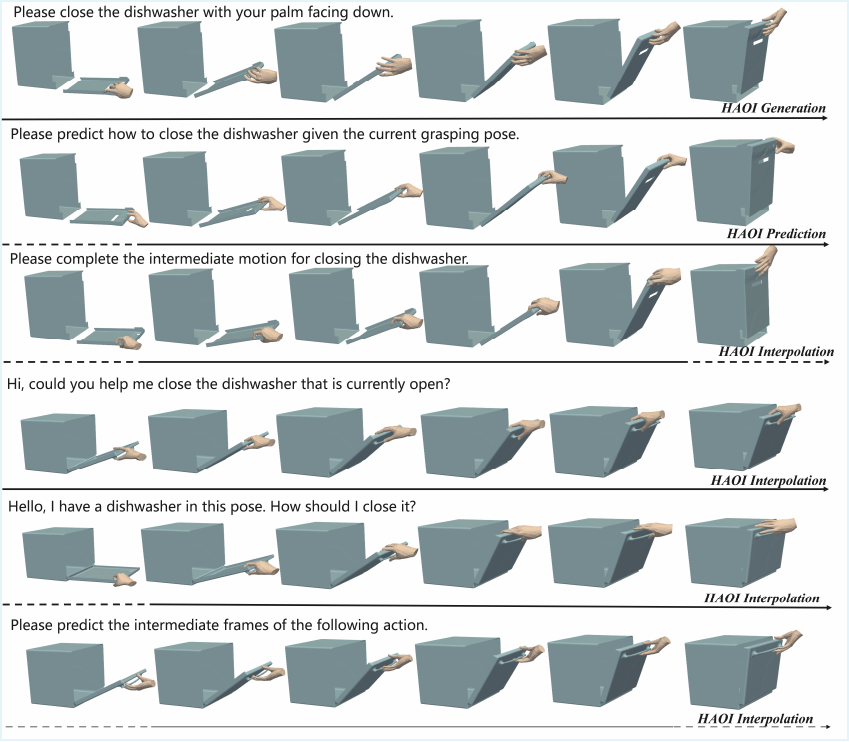}
  \caption{HAOI generation, prediction, and completion on large objects: dishwasher.}
  \label{fig:large_objects_2}
\end{figure*}

\begin{figure*}[htb]
  \centering
  \includegraphics[width=1\linewidth]{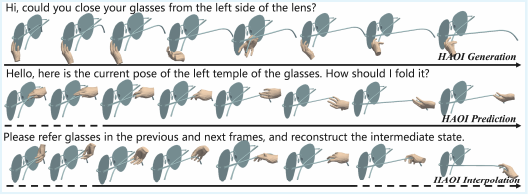}
  \caption{Representative failure case of our model.}
  \label{fig:failurecase}
\end{figure*}


\end{document}